\documentclass[twoside]{article}
\usepackage{amsmath}
\usepackage{amsfonts}
\usepackage{amssymb}
\usepackage{bm}
\usepackage{forest}
\usepackage{subfigure}
\usepackage{graphicx}
\usepackage{tabularx}
\usepackage{graphics}
\usepackage{multirow}
\usepackage{lipsum}
\usepackage{multicol}
\usepackage{url}
\usepackage{mdwlist}
\usepackage{booktabs}
\usepackage{aistats2016}

\begin{document}

%

%

\twocolumn[

\aistatstitle{Taxonomy grounded aggregation of classifiers with different label sets}


\aistatsauthor{ Amrita Saha \And Sathish Indurthi \And Shantanu Godbole \And Subendhu Rongali \And Vikas C. Raykar}
\vspace{1.5em}
\aistatsaddress{ IBM Research, India} ]

\begin{abstract}
 We describe the problem of aggregating the label predictions of diverse classifiers using a class taxonomy. Such a taxonomy may not have been available or referenced when the individual classifiers were designed and trained, yet mapping the output labels into the taxonomy is desirable to integrate the effort spent in training the constituent classifiers. A hierarchical taxonomy representing some domain knowledge may be different from, but partially mappable to, the label sets of the individual classifiers. We present a heuristic approach and a principled graphical model to aggregate the label predictions by grounding them into the available taxonomy. Our model aggregates the labels using the taxonomy structure as constraints to find the most likely hierarchically consistent class. We experimentally validate our proposed method on image and text classification tasks.
\end{abstract}

\section{Introduction}
In several real-world classification problems (for example visual object recognition~\cite{krizhevsky_NIPS_2012_imagenet}, text categorization~\cite{koller_ICML_1997,partalas2015lshtc}, web content classification~\cite{dumais_SIGIR_2000}, US Patent codes, ICD~\cite{ICD} codes of diseases~\cite{lita2008large} etc.) the classes to be predicted are naturally organized into a large pre-defined \emph{class hierarchy} or a \emph{class taxonomy}---typically a tree or a DAG (Directed Acyclic Graph). However most state-of-the-art results are obtained with \emph{flat classifiers} which typically ignore the class hierarchy and treat each class separately. It could be that the taxonomy was not available while training the classifiers or that it was not explicitly used. These hierarchy agnostic flat classifiers can be either multi-class, multi-label, or binary classifiers possibly trained on subsets of the classes. In this paper we address the problem of aggregating the output of multiple such flat classifiers using the pre-defined class taxonomy, since several applications need grounded references to such background taxonomies.

As a motivating example we consider the task of visual object recognition~\cite{krizhevsky_NIPS_2012_imagenet,girshick_CVPR_2014_RCNN}---given an image (or a region in the image) the task is to predict the most likely object in the image. Considerable amount of progress has been achieved in the computer vision community and various pre-trained state-of-the-art flat classifiers are available. Since the number of possible objects is quite large these classifiers are trained with different datasets and class labels\footnote{For example the CIFAR-100 dataset~\cite{CIFAR_dataset} has 100 class labels, the PASCAL-VOC datset~\cite{everingham2010pascal} has 20 class labels, and the latest ImageNet ILSVRC challenge dataset~\cite{ILSVRC15} has 1000 class labels.} (not necessarily mutually exclusive). Considerable amount of research effort and also CPU time(especially in the case of convolutional neural network based approaches~\cite{krizhevsky_NIPS_2012_imagenet,szegedy_2014_googlenet} which takes weeks to months to train) has been spent on training these classifiers. Hence we are interested in reusing multiple such classifiers on a given image, aggregating the scores and predicting the final label. However combining such pre-trained flat classifiers poses its own set of challenges which we list below and address in this paper.

\noindent \emph{\textbf{Different class labels}}---The classifiers are generally trained with different class labels based on the labeled dataset it was trained on. For example we may have one classifier trained with a label set $\{dog,cat,bird\}$ while another classifier could have been trained with the label set $\{doberman,airplane,animal\}$. We will use the class taxonomy in order to ground the different label sets into a common space. For this domain we use the  Wordnet~\cite{wordnet}~\footnote{WordNet~\cite{wordnet} is a large lexical database of english where nouns are grouped into sets of cognitive synonyms (synsets), each expressing a distinct concept (around 80k synsets). Synsets are interlinked by means of conceptual-semantic and lexical relations. The most frequently encoded relation among synsets is the super-subordinate relation (also called hyperonymy, hyponymy or \textsf{IS-A} relation). It links more general synsets like $\{dog\}$ to increasingly specific ones like $\{doberman\}$.}
, which is a DAG structured class taxonomy organized by the hypernym-hyponym hierarchy. The Wordnet encodes our prior world knowledge that $doberman$ is a $dog$ and $dog$ is an $animal$.

\noindent \emph{\textbf{Hierarchically inconsistent predictions}}---Since the flat classifiers are trained ignoring the taxonomy, the class predictions may not be hierarchically consistent. Consider, for example, a classifier trained with class labels $\{doberman,dog,cat\}$. Since the class labels are not mutually exclusive, it is possible that the classifier may give a high score for $doberman$ and a low score for $dog$, though the taxonomy implies that a $doberman$ is a $dog$ and hence $dog$ should also receive a high score. The same problem persists across different classifiers---if one classifier predicts an instance as a $dog$ and another as $doberman$, we need to aggregate the two classifiers in a hierarchically consistent way.

\noindent \emph{\textbf{Different classifier accuracies}}---The individual classifiers can have (unknown) different accuracies, which have to be accounted for when aggregating the classifiers. Moreover the reported accuracies are based on the classes on which it was trained on and also ignore the hierarchy. We model the classifier performance using the taxonomy and estimate the accuracies using a validation set.

\noindent \emph{\textbf{Depth of the predicted class label}}---By using the class taxonomy we can actually predict a class label which is not in the label sets used to train the classifiers. We generalize the notion of the label of an instance to a path in the taxonomy. The path starts at the root of the taxonomy and terminates at any class (not necessarily the leaf node) in the taxonomy. We propose strategies to decide where to terminate the path to give a final prediction.

\noindent \emph{\textbf{Related work}}---There is a rich literature in the area of hierarchical classification (see \cite{silla_DMKD_2011} for a survey), which deals with training classifiers by explicitly accounting for the class hierarchy. However in this paper we are primarily concerned with aggregating pre-trained flat classifiers in a hierarchically consistent way. We do not attempt to re-train any classifiers using the taxonomy.

The proposed algorithms can also be used to aggregate (hierarchical) labels collected from multiple annotators via crowdsourcing. While sophisticated techniques exist for binary, categorical and ordinal labels~\cite{raykar_JMLR_2010_crowds} to the best of our knowledge there are no methods for labels defined by a taxonomy.

Another area of research related to our problem setting is that of integrating (or mapping) label-sets into each other as in the case of e-commerce catalog integration \cite{agrawal_www01, sarawagi_kdd03, taci_tkde13}. In catalog integration the problem is mapping a source product taxonomy (of a seller) containing textual descriptions of products on to a target master taxonomy (of the ecommerce site). Various techniques for this involve jointly learning class mappings with or without data labeled with  both label-sets, and estimating and re-training the master taxonomy using statistics and/or structure of the source taxonomy. The specific formulation we propose is quite different from such a label-set mapping problem since we want to lazily combine only the predictions of differently trained classifiers into a background taxonomy. As in our motivating example of mapping object classification predictions to WordNet,  we want to re-use the effort spent in creating and tuning existing classifiers and view them through the lens of WordNet to naturally describe objects in images as per its world knowledge.

In our problem setting the constituent classifiers are free to evolve and change, and have their own area of expertise. Thus ensemble meta-learning methods that learn accuracy estimates or dynamic model selection techniques do not apply to our setting.

\noindent \emph{\textbf{Organization}}---\S~\ref{sec:statement} introduces the notation and the problem statement. We formally define the notion of class taxonomy and hierarchical consistency in a taxonomy. \S~\ref{sec:heuristic} presents a heuristic solution based on propagating scores in the class taxonomy and generalizes the notion of the label of an instance to a path in the taxonomy and specifies when to terminate the path. \S~\ref{sec:graphical_model} presents the proposed graphical model. In \S~\ref{sec:EM} we discuss some extensions and present an EM algorithm to estimate the parameters of the graphical model without having access to any validation set. \S~\ref{sec:exp} experimentally validates the various approaches on a visual object recognition and a text classification dataset.

\section{Notation and problem statement}\label{sec:statement}


\noindent \emph{\textbf{Class Taxonomies}}---In this paper we are concerned with classification problems where the class labels are organized hierarchically into class taxonomies. The hierarchy imposes a parent-child \textsf{IS-A} relation among the classes---an instance belonging in a specific class, also belongs in all its ancestor classes. Formally a \emph{class taxonomy}~\cite{silla_DMKD_2011} is defined as a pair $\left(\mathcal{C},\prec\right)$, where $\mathcal{C}=\left\{c_1,\hdots,c_k\right\}$ is a finite set of $k$ classes organized hierarchically with the \textsf{IS-A} relationship $\prec$. For any two classes $c_i$ and $c_j$ the relation $c_i \prec c_j$ means that $c_i$ is a sub-class of $c_j$. The relationship $\prec$ satisfies the following properties: (1) \emph{Asymmetry} If $c_i \prec c_j$ then $c_j \nprec c_i$, $\forall c_i,c_j\in\mathcal{C}$. (2) \emph{Anti-reflexivity} $c_i \nprec c_i$, $\forall c_i\in\mathcal{C}$. (3) \emph{Transitivity} If $c_i \prec c_j$ and $c_j \prec c_k$ then $c_i \prec c_k$, $\forall c_i,c_j,c_k\in\mathcal{C}$.
In graphs with cycles, only the transitivity property holds. In this paper we consider only hierarchies without cycles. The parent of a class $c \in \mathcal{C}$ is denoted as $\uparrow(c)$ and its child is denoted as $\downarrow(c)$. The descendants and ancestors are denoted $\Downarrow(c)$ and $\Uparrow(c)$ respectively.

The taxonomy can either be a \emph{tree} (where the classes have a single parent) or a \emph{directed acyclic graph}(DAG) (where the classes can have multiple parents). The Wordnet is a DAG taxonomy organized by the hypernym-hyponym hierarchy (See Figure~\ref{fig:wordnet_DAG} for an illustration).

{\footnotesize
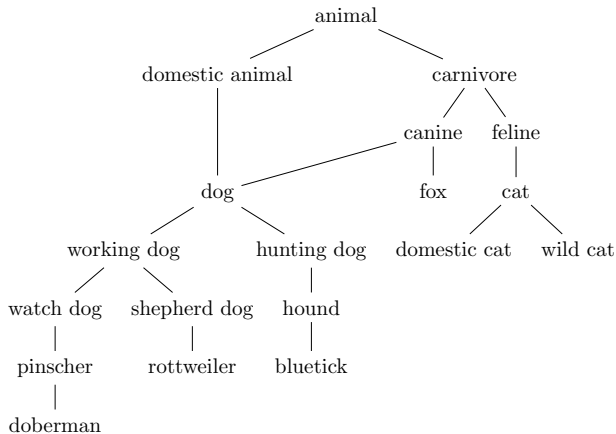
\begin{figure}
  \centering
  \scalebox{0.8}{
  \begin{forest}
[animal
    [domestic animal[dog,name=dog,tier=entry[working dog[watch dog[pinscher[doberman]]][shepherd dog[rottweiler]]][hunting dog[hound[bluetick]]]]]
    [carnivore,name=carnivore[canine,name=canine[fox,tier=entry]][feline[cat,tier=entry[domestic cat][wild cat]]]]
]
\draw[-] (dog) to (canine);
\end{forest}
}
  \caption{\small{A small subset of the Wordnet illustrating the \textsf{IS-A} DAG taxonomy. In this example $dog$ has two parents $domestic\:animal$ and $canine$.}}\label{fig:wordnet_DAG}
\end{figure}
}
\noindent \emph{\textbf{Flat pre-trained classifiers on taxonomy subsets}}---We have a set of $m$  trained classifiers $f^1,\hdots,f^m$. Each classifier $f^j$ is trained using a subset of $k^j \leq k$ classes $\mathcal{C}^j=\left\{c_1,\hdots,c_{k^j}\right\}$. We work with the assumption that these classifiers are flat multi-class classifiers~\footnote{The classifier can also be a binary classifier trained for a class $c$.} \emph{trained without the knowledge of the class hierarchy}. It is also possible that the same classifier may have a category for a parent class (for example \emph{dog}) and also a separate category for a descendant class as well (for example \emph{doberman}). Let $y^{j}(c)=f^j(\bm{x},c)$ be the score assigned to class $c$ to an instance $\bm{x}$ by the classifier $f^j$. Without loss of generality we assume that these scores are probabilities~\footnote{For real valued scores they can be converted to probabilities via the soft-max function or via some calibration techniques~\cite{zadrozny2002transforming}.}, that is, $y^{j}(c)=\text{Pr}(z(c)=1|\bm{x},f^j)$, where $z(c)$ is the true binary label of the instance for class $c$ in the taxonomy. We further note that the accuracies of the $m$ classifiers are different. In the graphical model presented later (\S~\ref{sec:graphical_model}) we estimate the accuracies via a validation set and later present an EM algorithm (\S~\ref{sec:EM}) to estimate them without using a validation set.

We implicity assume that the classifier label sets are mappable to some classes in the taxonomy. If this is not true then we can approximate the mappings using class mapping techniques~\cite{agrawal_www01,sarawagi_kdd03,taci_tkde13}.

\noindent \emph{\textbf{Generalization of the instance label to paths in the taxonomy}}---Given a taxonomy the true label of any instance $\bm{x}$ is completely specified by one of the classes in the leaf nodes of the hierarchy. However in practice the class label may be specified by any class of the taxonomy higher than the leaf node. For example, the true label for an instance $\bm{x}$ maybe $doberman$ (which is a leaf node in the Wordnet taxonomy), however the class label as specified by an annotator (or the classifier) could be one of its hypernyms, for example, $dog$.

We generalize the notion of the label of an instance $\bm{x}$ to a \emph{label path} in the taxonomy. The path starts at the root of the taxonomy and terminates at any class in the taxonomy.  Paths that end at the leaf node are completely specified. For any instance $\bm{x}$ with class label $c$ (not necessarily the leaf node) we denote $\text{path}(c)=\left\{c,\uparrow(c),\uparrow(\uparrow(c)),\hdots,\right\}$  as the set of all classes starting from $c$ to the root node. For a tree taxonomy there is one unique path from the terminal node $c$ to the root. For a DAG taxonomy there could be multiple such paths~\footnote{For example, in the Wordnet hierarchy the following two hypernym paths can be found for the class $doberman$ since $dog$ has two parents $domestic\:animal$  and $canine$.

$\text{path}_{1}({doberman})=$ [doberman, pinscher, watchdog, working dog, dog, domestic animal, animal, organism,living thing, whole, object, physical entity, entity]

$\text{path}_{2}({doberman})=$ [doberman, pinscher, watchdog, working dog, dog, canine, carnivore, placental, mammal, vertebrate, chordate, animal, organism, living thing, whole, object, physical entity, entity]}.

\noindent \emph{\textbf{Taxonomy consistency}}---Since the classifiers are trained without the knowledge of the taxonomy, it is very likely that the scores may not be hierarchically consistent. For example, an instance may possibly have a higher score for $doberman$ than $dog$. However from the taxonomy we know that a $doberman$ is a $dog$, hence $dog$ should have as high a score as $doberman$. We define the following notion of \emph{taxonomy consistency}: For an instance $\bm{x}$ with true class label $c$, that is if $z(c)=1$, taxonomy consistency implies $z(c_j)=1$ for all $c_j \in \Uparrow(c)$, where $\Uparrow(c)$ are ancestors of $c$. This implies that $\text{Pr}[z(c_j)=1|z(c)=1]=1$  for all $c_j \in \Uparrow(c)$.

\noindent \emph{\textbf{Problem statement}}---Given a set of $n$ instances $\bm{x}_1,\hdots,\bm{x}_n$, a taxonomy $\left(\mathcal{C},\prec\right)$, where $\mathcal{C}=\left\{c_1,\hdots,c_k\right\}$ is a set of $k$ classes organized hierarchically with the \textsf{IS-A} relationship $\prec$, a set of $m$ trained classifiers $f^1,\hdots,f^m$, each classifier $f^j$ is trained using a subset of $k^j \leq k$ classes $\mathcal{C}^j=\left\{c_1,\hdots,c_{k^j}\right\}$, and $y^{j}_{i}(c)$ be the score assigned to class $c\in\mathcal{C}^j$ for the instance $\bm{x}_i$ by the classifier $f^j$,
the task is to aggregate the scores and estimate the best path(s) in the taxonomy $\text{path}_{i}(c)$ for every instance $\bm{x}_i$ such that the true class label $c$ is taxonomically consistent. We could have multiple paths and each path need not necessarily end at a leaf node~\footnote{This is the case when we do not have enough confidence to make a prediction till the leaf node. In such case we backoff to an ancestor.}.

\section{Score propagation in the class taxonomy}\label{sec:heuristic}
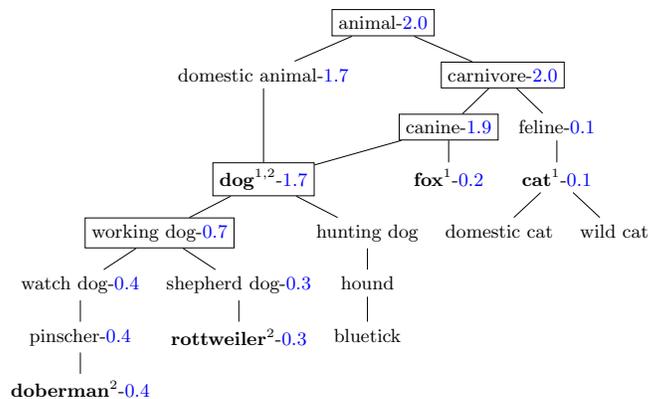
\begin{figure}[!ht]
\centering
{\footnotesize
  \scalebox{0.8}{
  \begin{forest}
    [animal-\small{\textcolor[rgb]{0.00,0.00,1.00}{2.0}},draw
    [domestic animal-\small{\textcolor[rgb]{0.00,0.00,1.00}{1.7}}[\textbf{dog}$^{1,2}$-\small{\textcolor[rgb]{0.00,0.00,1.00}{1.7}},name=dog,tier=entry,draw
    [working dog-\small{\textcolor[rgb]{0.00,0.00,1.00}{0.7}},draw
    [watch dog-\small{\textcolor[rgb]{0.00,0.00,1.00}{0.4}}
    [pinscher-\small{\textcolor[rgb]{0.00,0.00,1.00}{0.4}}
    [\textbf{doberman}$^2$-\small{\textcolor[rgb]{0.00,0.00,1.00}{0.4}}
    ]]]
    [shepherd dog-\small{\textcolor[rgb]{0.00,0.00,1.00}{0.3}}
    [\textbf{rottweiler}$^2$-\small{\textcolor[rgb]{0.00,0.00,1.00}{0.3}}
    ]]]
    [hunting dog
    [hound
    [bluetick
    ]]]]]
    [carnivore-\small{\textcolor[rgb]{0.00,0.00,1.00}{2.0}},name=carnivore,draw
    [canine-\small{\textcolor[rgb]{0.00,0.00,1.00}{1.9}},name=canine,draw
    [\textbf{fox}$^1$-\small{\textcolor[rgb]{0.00,0.00,1.00}{0.2}},tier=entry
    ]]
    [feline-\small{\textcolor[rgb]{0.00,0.00,1.00}{0.1}}
    [\textbf{cat}$^1$-\small{\textcolor[rgb]{0.00,0.00,1.00}{0.1}},tier=entry
    [domestic cat
    ]
    [wild cat
    ]]]]
]
\draw[-] (dog) to (canine);
\end{forest}
 }
 \caption{\small{\emph{Illustration of the heuristic score propagation algorithm (\S~\ref{sec:heuristic})} We have two different classifiers $f^1$ (trained with class labels $\{dog,fox,cat\}$) and $f^2$ (trained with class labels $\{dog,doberman,rottweiler\}$). The classifiers assign the following scores to an test instance $\bm{x}$: $y^1(dog)=0.7$, $y^1(fox)=0.2$, $y^1(cat)=0.1$ and $y^2(dog)=0.3$, $y^2(doberman)=0.4$, $y^2(rottweiler)=0.3$. The final scores for each class after propagating the scores to the ancestors and summing them up are also shown. The final predicted label path for the instance is marked by rectangles. Note that the path terminates into a non-leaf class $working\:dog$.}}\label{fig:heuristic_illustration}}
\end{figure}

We will first present a heuristic solution by propagating the classifier scores upward from a particular class to all its ancestors in the taxonomy by navigating the IS-A hierarchy upwards. The scores from multiple classifiers at a node are then aggregated by summing them up. The final path is then estimated by traversing the taxonomy from the root and terminating at a class based on the entropy of the children. Specifically (See Figure~\ref{fig:heuristic_illustration} for an illustration)
\begin{itemize}
\item First construct an induced sub-graph with classes $C_I=\{C^1 \cup \hdots \cup C^m\}$  (which is the union of all classes from the $m$ flat classifiers $f^1,\hdots,f^m$) and all its ancestors.
\item Initialize the scores $p(c)$ of each of the class-nodes $c$ in the graph to $0$.
\item For each classifier $f^j$, add the scores $y^j(c)~\forall~c~\in~{C^j}$ to $p(c')$ where $c' \in {\Uparrow(c) \cup c}$.
\item {Initialize path to null. Starting with the root node, recursively perform the following steps for every node in the path being constructed.
    \begin{itemize}
    \item Add node $c$ to path.
    \item Let $Y_c$ be the set of scores $p(c')~\forall~c'~\in\downarrow(c)$.
    \item Calculate the entropy $g_c$ of the set $Y_c$ and normalize it based on the number of children.
    \item if $g_c > \theta$ (a pre defined threshold) or $\downarrow(c)$ is null, quit.
    \item else select the node in $\downarrow(c)$ with the highest value of score $p$.
    \end{itemize}}
\end{itemize}
By calculating the entropy at a particular node, we are trying to check if we can go further down the taxonomy for the instance. If the children have equal scores (if the impurity exceeds a threshold value $\theta$.) then we do have have enough evidence to make a decision. Hence we back-off and terminate the path at this node.

\section{The proposed probabilistic graphical model for aggregating classifiers}\label{sec:graphical_model}
\begin{figure}[!h]
  \includegraphics[height=3.2in,width=0.8\linewidth]{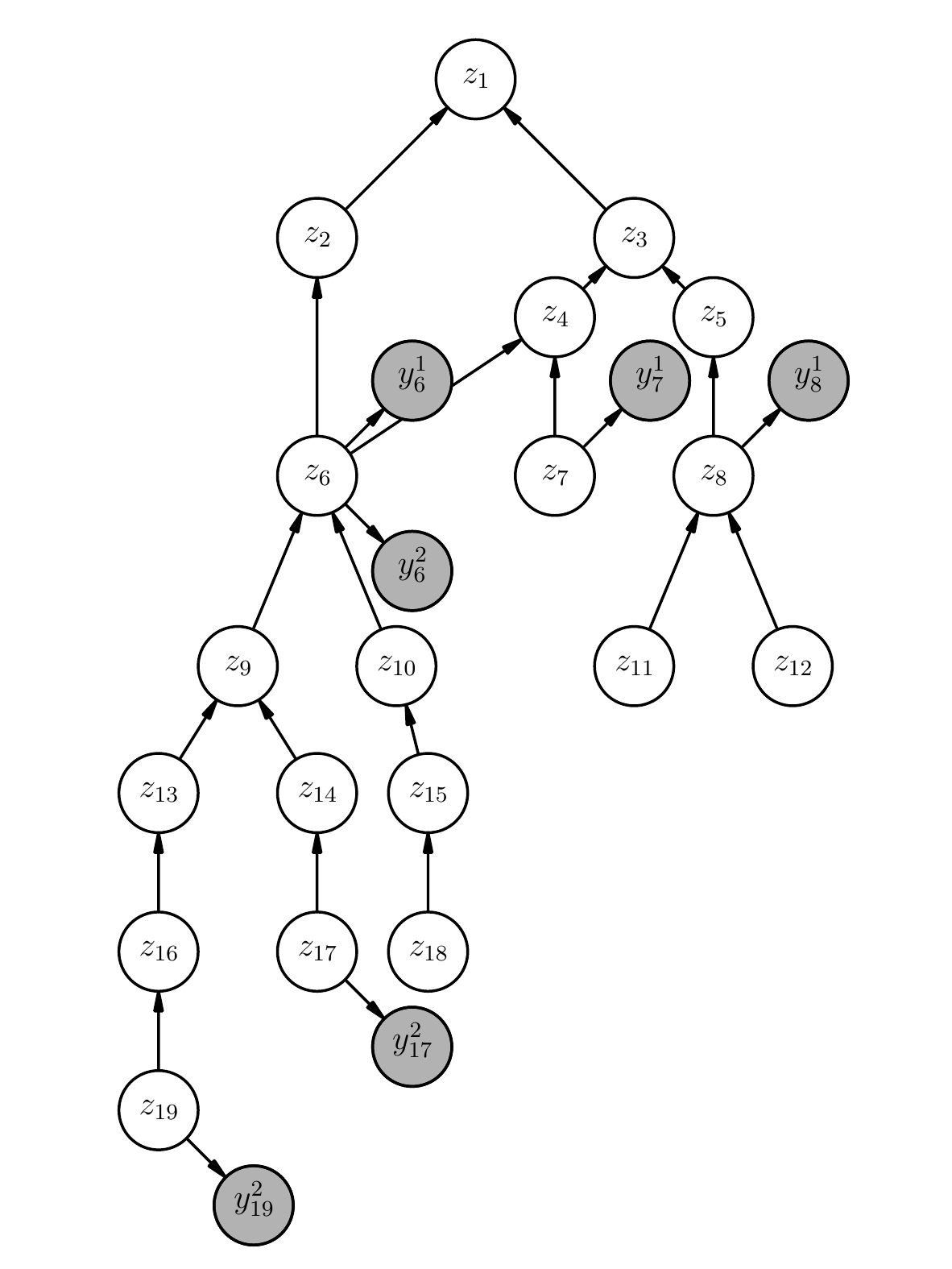}
  \vspace{0.5pt}
  \\
  \\
  \caption{\small{The proposed graphical model (\S~\ref{sec:exp}) for the two classifier label aggregation problem illustrated in Figure~\ref{fig:heuristic_illustration}. The solid $y$ nodes (continuous) correspond to the classifier predictions and the other $z$ nodes (discrete) correspond the classes in the taxonomy.}}\label{fig:graphical_model}
\end{figure}

The heuristic method assumes that all the classifiers have the same performance and then aggregates the scores. In this section we cast the label aggregation problem as an inference problem in an appropriately defined graphical model (for a given instance $\bm{x}$). The proposed graphical model (or Bayesian network) has two kinds of nodes, discrete binary nodes corresponding to hierarchically organized classes in the taxonomy and continuous nodes corresponding to the $m$ classifier scores (see Figure~\ref{fig:graphical_model}, which is the graphical model corresponding to the induced sub-graph in Figure~\ref{fig:heuristic_illustration}).

Each class $c$ in the taxonomy corresponds to a binary discrete node $z(c)$ which is the true (unknown) binary label of the instance $\bm{x}$ for class $c$ in the taxonomy. All the discrete nodes $z_1,\hdots,z_K$ are organized hierarchically with the \textsf{IS-A} relationship $\prec$ defined by the class taxonomy---which defines the conditional independence assumptions of the graphical model. Each binary node $z_k$ is conditioned on its child $\downarrow(z_k)$ node. Taxonomy consistency is ensured by appropriately setting the conditional probability distribution as follows:
\begin{eqnarray}
  \text{Pr}[z_k=1|ch(z_k)]&=&1\:\text{if any}\:\downarrow(z_k)=1,
\end{eqnarray}
where $ch(z_k)$ are all the children of $z_k$. The other entries $\text{Pr}[z_k=1|ch(z_k)]$ are estimated using the structure of the taxonomy~\footnote{This is a reasonable estimate in the absence of any other information and works well in practice. If the actual counts from a corpus are available (for example the Brown corpus for WordNet) for each class in the taxonomy we can get more precise estimates.}.
 \vspace{-0.5pt}

For a classifier $f^{j}$ and a class $c_k \in \mathcal{C}^j$ (which is in the classifier class set) we have a continuous node $y^j_k$ which is conditioned on the discrete node $z_k$.We assume that conditioned on the true label $z_k$ the classifier score $y^j_k$ is normally distributed~\footnote{This is a reasonable model for scores constructed as a linear (or non-linear) combination of many features. For probabilistic classifiers since the scores lie in the range $[0,1]$ we first apply a logit or the inverse softmax transformation to the scores which makes the scores approximately normal.}
, that is,
\begin{eqnarray}
\centering
y^j_k|z_k=1&\sim&\mathcal{N}\left(y^j_k|\mu^j_{1k},(\sigma^j_{1k})^2\right)
\end{eqnarray}
\begin{eqnarray}
y^j_k|z_k=0&\sim&\mathcal{N}\left(y^j_k|\mu^j_{0k},(\sigma^j_{0k})^2\right)
\end{eqnarray}
where $\mathcal{N}\left(x|\mu,\sigma^2\right)$ is a normal distribution with mean $\mu$ and variance $\sigma^2$.  The model parameters $\bm{\theta}_k^j=\left\{\mu^j_{0k},\sigma^j_{0k},\mu^j_{1k},\sigma^j_{1k}\right\}$ are estimated by using a validation set with known class labels. We further assume that conditioned on the true label $z_k$ the $m$ classifiers are independent.

Having specified the graphical model and the model parameters the task is to find the most likely configuration of the class nodes $z_k$ given the values for the classifier nodes $y^j_k$. We actually compute the marginal probability for each of the class nodes and then traverse the taxonomy using the method described in \S~\ref{sec:heuristic}. The graphical model which we have described is a \emph{mixed discrete-gaussian network}~\cite{cowell2006probabilistic}. A mixed discrete-gaussian network consists of both discrete and continuous nodes. The models assumes that the conditional distribution of the continuous variables, given the discrete, is multivariate Gaussian.  For such networks exact inference algorithms exist~\cite{lauritzen1992propagation}, which permits the local computation on the junction tree of exact probabilities, means and variances.


\section{Extensions}\label{sec:EM}

\noindent \emph{\textbf{Discrete labels}}---We can also incorporate classifiers which produce discrete labels. Instead of the bi-normal distribution we have the following two parameters which define the conditional probability distribution at each classifier node: $\alpha^j_k := \text{Pr}[y_k^j=1|z_k=1]$ and $\beta^j_k := \text{Pr}[y_k^j=0|z_k=0]$.

\noindent \emph{\textbf{The EM algorithm for estimating classifier parameters}}---In \S~\ref{fig:graphical_model} we estimated the parameters by using a separate validation set. Sometimes we may not have access to a labeled validation set. This is especially true when we are interested in aggregating the crowdsourced labels where the goal is estimate the true labels. In such scenarios we can estimate the model parameters directly via the Expectation Maximization(EM) algorithm. The EM algorithm~\cite{dempster_JRSS_1977} is an efficient iterative procedure to estimate the parameters in presence of missing/hidden data. We will use the unknown hidden true label $[z_1,\hdots,z_K]$ as the missing data in our case. Each iteration of the algorithm consists of two steps: an Expectation(E)-step and a Maximization(M)-step. The M-step involves maximization of a lower bound on the log-likelihood that is refined by the E-step. Specifically, in the E-step we obtain the marginal probabilities of all the class nodes given the current estimate of the model parameters and the observed classifier scores. In the M-step we re-estimate the model parameters given the class labels (marginal probabilities) for the taxonomy class nodes. The only difference being that when estimating the parameters of the bi-normal model we need to use these marginal probabilities instead of the  binary labels (from the validation set). These two steps (the E- and the M-step) can be iterated till convergence.

\noindent \emph{\textbf{Entry level categories}}---The proposed algorithm returns a path in the taxonomy. The final label is computed by terminating the path based on the entropy. However for some domains we may want to terminate based on other domain specific criterion. For example, in object recognition task we may want to assign the label of the object to an `entry level' category--the labels people will use to name an object, for example, the entry level category for $rottweiler$ is $dog$. We can appropriately modify our termination strategy to account for this by suitable backing off the path using ideas in \cite{Ordonez:2013:entrylevel}.

\section{Experimental Validation}\label{sec:exp}
We experimentally validate our proposed algorithms on two domains, object recognition in images and text categorization. In both these domains we have a natural pre-defined taxonomy.

\noindent \emph{\textbf{Datasets}}---For visual object recognition we use a subset of images from the ImageNet ILSVRC2014 detection challenge dataset~\cite{ILSVRC15}. The dataset consists of the 200 basic categories that map to a total of 547 WordNet synsets (including the children). Out of this 547 node taxonomy we extract a sub-hierarchy rooted at the basic category $animal$ resulting in a taxonomy of 297 nodes. The set of images under these 297 categories are completely non-overlapping and the aggregated set of 70096 images forms the final dataset for our experiments. For our text categorization experiments, we used the benchmark Reuters Corpus Volume 1 (RCV1) news articles dataset~\cite{rcv}. This is a hierarchical multi-labeled dataset where over 138 thousand articles are tagged along topics, industries, and regions facets. We used the 354 class industries taxonomy in our experiments.

\noindent \emph{\textbf{Experimental setup}}---We use $60\%$ of the data for training, $10\%$ for validation and $30\%$ for testing. We randomly select a subset of classes from the taxonomy and train a multi-class flat classifier by completely ignoring the class taxonomy. For object detection, we take the activations of the sixth  hidden layer of a deep convolutional neural network as features and then train a linear multi-class SVM with these features~\cite{decaf}. For the text categorization we used the standard token pre-processing available\footnote{\url{http://www.jmlr.org/papers/volume5/lewis04a/lyrl2004_rcv1v2_README.htm}} for this dataset with tf-idf representation and trained a linear SVM. The classifier is trained using the train split with the validation split being used to tune any hyper-parameters. We experimented with $10$ flat classifiers each operating with different(with some overlap) class labels (see Table~\ref{tab:results} for the number of classes). For each of the flat classifiers, if a training sample belongs to multiple labels (where one label is a parent of another) then during training-set construction, we randomly assign the instance to one of the classes in order to avoid confounding the classifier. The final model performance is evaluated on the test split which has a representation of all the classes in the taxonomy.

\noindent \emph{\textbf{Evaluation metrics}}---Accuracy is not a good metric for evaluating the performance of the various classifiers because of the hierarchical relations among classes. The classifiers operate only on a subset of the labels in the taxonomy, but the test set can possibly have instances from all the classes in the taxonomy. There is a rich literature on evaluation measures for hierarchical classification (see \cite{kosmopoulos:2015:evaluation} for a review). For our evaluation we choose the Lowest Common Ancestor (LCA) based precision $(P_{LCA})$, recall $(R_{LCA})$ and $F_1$-score $(F_{LCA})$ measures as recommended in \cite{kosmopoulos:2015:evaluation}. These measures are essentially hierarchical versions of the precision, recall, and $F_1$-score based on LCA of the actual and the predicted class.

\noindent \emph{\textbf{Results}}---Table~\ref{tab:results} shows the mean (and the standard deviation) hierarchical precision $(P_{LCA})$, recall $(R_{LCA})$ and $F_1$-score $(F_{LCA})$ on the test set for each of the individual classifiers and also our proposed heuristic(\S~\ref{sec:heuristic}) and the graphical model(\S~\ref{sec:graphical_model}) based aggregation algorithms for both the visual object recognition and the text classification tasks. For the graphical model inference we used the junction tree algorithm from the Bayes Net Toolbox\footnote{\url{https://code.google.com/p/bnt/}}. The appropriate thresholds for deciding the terminal class were tuned using the validation set. For the heuristic algorithm the path termination was based on the entropy. For the graphical model the terminal node was decided based on the marginal probabilities. For the visual object recognition dataset on all the three measures the performance of the aggregation algorithms are better than or equal to the best performing classifier in the ensemble, the graphical model based approach outperforming the heuristic approach. For the object recognition dataset we also show results for an alternate termination strategy described in \S~\ref{sec:EM} which decides the terminal node by backing off to a suitable entry level class~\cite{Ordonez:2013:entrylevel}. This alternate termination strategy gave further improvements. For the text classification dataset the graphical model outperforms the best classifier in terms of the precision by a large margin.

\begin{table}[h!]
\small{
\setlength\tabcolsep{5.3pt}
\setlength{\extrarowheight}{0.1cm}
\begin{tabular}{|p{1.55cm}|p{0.75cm}|c|c|c|c|@{}}
\multicolumn{5}{l}{\textbf{~~~visual object recognition:} ~~ 18246 instances } \\\hline
Model &classes  & $P_{LCA}$ & $R_{LCA}$ & $F_{LCA}$\\ \hline
classifier0 &16& 0.35 [0.27] & \underline{0.37} [0.27] & \underline{0.35} [0.26] \\
classifier1 &10& 0.33 [0.21] & 0.35 [0.22] & 0.33 [0.21] \\
classifier2 &18& \underline{0.36} [0.28] & 0.27 [0.22] & 0.30 [0.21] \\
classifier3 &14& 0.32 [0.20] & 0.33 [0.16] & 0.31 [0.14] \\
classifier4 &15& 0.32 [0.14] & 0.31 [0.16] & 0.31 [0.14] \\
classifier5 &10& 0.31 [0.14] & 0.31 [0.16] & 0.30 [0.13] \\
classifier6 &16& 0.24 [0.11] & 0.31 [0.14] & 0.26 [0.11] \\
classifier7 &11& 0.21 [0.14] & 0.20 [0.13] & 0.20 [0.13] \\
classifier8 &10& 0.23 [0.13] & 0.31 [0.16] & 0.26 [0.13] \\
classifier9 &14& 0.24 [0.12] & 0.33 [0.15] & 0.27 [0.12] \\[0.1cm]\hline &&&&\\[-0.3cm]
heuristic & & \textbf{0.40} [0.19] & \textbf{0.38} [0.15] & \textbf{0.38} [0.13]  \\[0.1cm] \hline&&&&\\[-0.3cm]
graphical & & \textbf{0.44} [0.21] & \textbf{0.39} [0.13] & \textbf{0.41} [0.13] \\[0.1cm] \hline
heuristic  &&&& \\[-0.1cm]
entry-level & & \textbf{0.57} [0.29] & 0.37 [0.12] & \textbf{0.44} [0.13] \\[-0.1cm]
termination &&&& \\[0.1cm]\hline
graphical &&&&\\[-0.1cm]
entry-level & & \textbf{0.58} [0.29] & \textbf{0.38} [0.12] & \textbf{0.46} [0.13] \\[-0.1cm]
termination &&&&\\[0.1cm]\hline\hline
\end{tabular}
}%
\end{table}
\begin{table}[h!]
\small{
\setlength\tabcolsep{5.5pt}
\setlength{\extrarowheight}{0.1cm}
\begin{tabular}{@{}|r|c|c|c|c|@{}}
\multicolumn{5}{l}{\textbf{~~~~text classification:} ~~ 41395 instances } \\\hline
Model &classes  & $P_{LCA}$ & $R_{LCA}$ & $F_{LCA}$\\ \hline 
classifier0 &10& 0.29 [0.12] & 0.32 [0.15] & 0.30 [0.13] \\
classifier1 &10& 0.36 [0.19] & 0.35 [0.20] & 0.35 [0.19] \\
classifier2 &10& 0.49 [0.27] & 0.54 [0.33] & 0.51 [0.28] \\
classifier3 &10& 0.52 [0.36] & 0.50 [0.34] & 0.52 [0.34] \\
classifier4 &10& 0.50 [0.28] & 0.44 [0.31] & 0.47 [0.29] \\
classifier5 &10& 0.41 [0.25] & 0.40 [0.26] & 0.40 [0.25] \\
classifier6 &10& 0.46 [0.32] & 0.49 [0.31] & 0.47 [0.30] \\
classifier7 &10& 0.54 [0.34] & \underline{0.55} [0.33] & \underline{0.54} [0.33] \\
classifier8 &10& \underline{0.55} [0.33] & 0.54 [0.34] & 0.53 [0.32] \\
classifier9 &10& 0.34 [0.15] & 0.32 [0.15] & 0.33 [0.15]  \\[0.1cm]\hline &&&&\\[-0.3cm]
heuristic & & 0.49 [0.27] & 0.46 [0.30] & 0.47 [0.28]  \\[0.1cm] \hline&&&&\\[-0.3cm]
graphical & & \textbf{0.77} [0.27] & 0.45 [0.28] & \textbf{0.56} [0.24]\\[0.1cm] \hline\hline
\end{tabular}
}
\caption{\small{\emph{Experimental results}(\S~\ref{sec:exp}) The mean (and the standard deviation) hierarchical precision $(P_{LCA})$, recall $(R_{LCA})$ and $F_1$-score $(F_{LCA})$  measures on the test set for each of the individual classifiers and also our proposed heuristic(\S~\ref{sec:heuristic}) and the graphical model(\S~\ref{sec:graphical_model}) based aggregation algorithms for both the visual object recognition and the text classification task. For each task and measure the best performing classifier is underlined and the proposed algorithms is in bold it the performance is better than the best individual classifier.}\vspace{-10pt}}\label{tab:results}
\end{table}

\section{Conclusion}
In this paper we formulated the problem of aggregating labels from multiple flat classifiers into a possible different hierarchical taxonomy for the labels. This was achieved without modifying or retraining the constituent classifiers in any way. We proposed two solutions, one based on a heuristic score propagation through the taxonomy and a more principled approach using a graphical model. The proposed algorithms were experimentally validated on two real world problems of visual object recognition and text categorization. We plan to extend this approach to taxonomies with cycles and curated knowledge graphs or ontologies. Our model implicity assumes that the classifier label sets are mappable to some classes in the taxonomy. We plan to integrate ideas from catalog integration~\cite{agrawal_www01} directly into our graphical model.

\small
\bibliographystyle{plain}
\bibliography{taxonomy,crowdsourcing}

\end{document}